\documentclass[letterpaper]{article} 
\usepackage{aaai25}  
\usepackage{times}  
\usepackage{helvet}  
\usepackage{courier}  
\usepackage[hyphens]{url}  
\usepackage{graphicx} 
\usepackage{natbib}  
\usepackage{caption} 
\usepackage{algorithm}
\usepackage{algorithmic}

\usepackage{subfigure}       
\usepackage{newfloat}
\usepackage{listings}
\DeclareCaptionStyle{ruled}{labelfont=normalfont,labelsep=colon,strut=off} 
\floatstyle{ruled}
\newfloat{listing}{tb}{lst}{}
\floatname{listing}{Listing}
\title{Everywhere Attack: Attacking Locally and Globally to Boost Targeted Transferability}
\author{
    Hui Zeng\textsuperscript{\rm 1,2}\equalcontrib,
    Sanshuai Cui\textsuperscript{\rm 3}\equalcontrib,
    Biwei Chen\textsuperscript{\rm 4}\thanks{Corresponding author.},
    Anjie Peng\textsuperscript{\rm 1}
}
\affiliations{
    \textsuperscript{\rm 1}Southwest university of science and technology, Mianyang, China
    \textsuperscript{\rm 2}Guangan institute of technology, Guangan, China\\
    \textsuperscript{\rm 3}City University of Macau, Macau, China
    \textsuperscript{\rm 4}Beijing normal university, Zhuhai, China
    
    zengh5@mail2.sysu.edu.cn, sanshuaicui@cityu.edu.mo,\\ 
    bchen@bnu.edu.cn, penganjie200012@163.com,
}

\usepackage{bibentry}

\begin{document}

\maketitle

\begin{abstract}
Adversarial examples' (AE) transferability refers to the phenomenon that AEs crafted with one surrogate model can also fool other models. Notwithstanding remarkable progress in untargeted transferability, its targeted counterpart remains challenging. This paper proposes an \textit{everywhere} scheme to boost targeted transferability. Our idea is to attack a victim image both globally and locally. We aim to optimize `an army of targets' in every local image region instead of the previous works that optimize a high-confidence target in the image. Specifically, we split a victim image into non-overlap blocks and jointly mount a targeted attack on each block. Such a strategy mitigates transfer failures caused by attention inconsistency between surrogate and victim models and thus results in stronger transferability. Our approach is method-agnostic, which means it can be easily combined with existing transferable attacks for even higher transferability. Extensive experiments on ImageNet demonstrate that the proposed approach universally improves the state-of-the-art targeted attacks by a clear margin, e.g., the transferability of the widely adopted Logit attack can be improved by 28.8\%$\sim$300\%. We also evaluate the crafted AEs on a real-world platform: Google Cloud Vision. Results further support the superiority of the proposed method.
\end{abstract}

%
 \begin{links}
     \link{Code}{https://github.com/zengh5/Everywhere_Attack}
 \end{links}

\section{Introduction}

Adversarial example (AE) \cite{sz:14} is a powerful tool for uncovering potential vulnerability of deep neural networks (DNN) before their deployment in security-sensitive applications \cite{ma:18}. An exciting property of the AE is that AEs crafted against one model have a non-negligible chance to fool unseen victim models, a.k.a., transferability. Numerous transferable attacks have emerged recently, e.g., stabilizing the optimization direction (Dong et al. 2018; Lin et al. 2020; Wan, Ye, and Huang 2021) or diversifying inputs and surrogates \cite{xie:19, d:19, wangb:21, lib:20, f:23}.

Despite extensive studies constantly refreshing transferability under the untargeted mode, targeted transferability is much more daunting since it requires unknown models outputting a specific label \cite{liu:17}. To bridge the gulf, tailored schemes for improving the transferability of targeted attacks have been proposed. For instance, resource-intensive attacks seek extra, target-specific classifiers \cite{i:20} or generators \cite{n:21, y:22} to optimize adversarial perturbations. Other researchers find that integrating novel loss functions with conventional simple iterative attacks can also enhance targeted transferability \cite{lia:20, zh:21, ze:23, Weng:23}.
\begin{figure}[t]
	\centering
	\includegraphics[width=0.95\columnwidth]{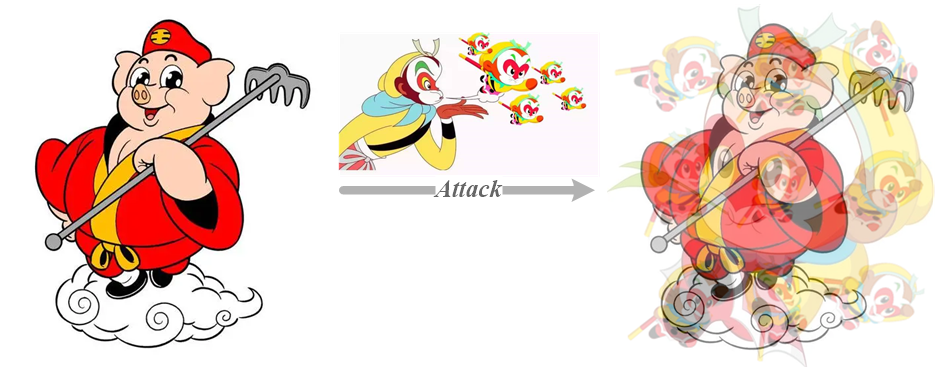} 
	\caption{Illustration of the proposed \textit{everywhere} attack. We attempt to synthesize an army of Wukongs (target, the monkey) into every local region of Bajie (victim, the pig).}
	\label{fig1}
\end{figure}

Despite the recent progress of targeted attacks, the reported transferability is still unsatisfactory. Unlike the attention regions (to the ground truth class) that are critical to untargeted attacks, which tend to overlap among diverse models  \cite{wu:20}, the target class-related attention regions vary significantly across different models (refer to Figure 2), resulting in limited targeted transferability. This paper proposes an \textit{everywhere} scheme to alleviate the attention mismatch dilemma for targeted attacks. Our idea is illustrated in Figure 1: Bajie (the pig) is expected to be attacked as Wu-Kong (the monkey)\footnote{Wukong and Bajie are the two main characters in the Chinese classical novel “Journey to the West.” Wukong can use his own hair to transform into a large number of clones, thus gaining an advantage in fighting.}. In contrast to conventional attacks that try to plant a high-confidence Wukong into the victim image, the proposed \textit{everywhere} attack simultaneously plants an army of Wukong in every local region of the victim image, with the hope that at least one Wukong falls into the attention area of the victim model. Our contributions can be summarized as follows.
\begin{itemize}
\item We note that a common cause of targeted transfer failure is the attention mismatch between the surrogate and victim models.
\item With this challenge in mind, we propose an \textit{everywhere} attack that tries to cover as much as possible the attention areas of various victim models. To our knowledge, this is the first attempt to enhance transferability by increasing the number of target objects, as opposed to previous works that aim to increase the confidence of the target class object.
\item Extensive experiments demonstrate that the proposed method possesses good extensibility and can improve almost all state-of-the-art targeted attacks by a clear margin.
\end{itemize}

\section{Related Work}
An adversarial attack typically has two modes: targeted and untargeted. A targeted attack misguides a classification model to produce an adversary-desired label, whereas an untargeted attack only fools it for misclassification. Targeted attacks are strictly more difficult yet pose a more severe threat to the classification model. In this section, we briefly review conventional tricks to improve untargeted transferability and then discuss tailored schemes for targeted transferability.

\subsection{Transferable Untargeted Attacks}

A plethora of transferable attacks is built up on the well-known iterative fast gradient sign method (IFGSM) \cite{k:16}, which can be formulated as:
\begin{equation}
	\textit{\textbf{I}}_{n+1}'=Clip_{\textit{\textbf{I}},\epsilon}(\textit{\textbf{I}}_n'+\alpha sign(\nabla_{\textit{\textbf{I}}_n'}J(\textit{\textbf{I}}_n', y_o)))
\end{equation}
\noindent where $\textit{\textbf{I}}_0'=\textit{\textbf{I}}$, $\nabla_{\textit{\textbf{I}}_n'}J()$ denotes the gradient of the loss function $J()$ with respect to $\textit{\textbf{I}}_n'$, $y_o$ is the original label, and $\epsilon$ is the perturbation budget. Researchers have proposed a variety of improved algorithms over IFGSM, e.g., the momentum iterative method (MI) \cite{d:18} integrates a momentum term into the iterative process. Diverse inputs method (DI) \cite{xie:19} and translation-invariant method (TI) \cite{d:19} leverage data augmentation to prevent attacks from overfitting a specific source model. Moreover, these enhanced schemes can be integrated for better transferability, e.g., Translation Invariant Momentum Diverse Inputs IFGSM (TMDI).

\subsection{Transferable Targeted Attacks}

In addition to the difficulties untargeted attacks face, targeted attacks have their own challenges, such as gradient vanishing \cite{lia:20, zh:21} and the restoring effect \cite{lia:20, ze:23}. Hence, tailored considerations are necessary for transferable targeted attacks. Existing efforts to boost targeted transferability can be divided into two families: resource-intensive methods and simple-gradient methods.

\textbf{Resource-intensive attacks} require training auxiliary target-class-specific classifiers or generators on additional data. In the feature distribution attack \cite{i:20}, a light-weight, one-versus-all classifier is trained for each target class $y_t$ at each specific layer to predict the probability that a feature map is from $y_t$. Transferable targeted perturbation (TTP) \cite{n:21} trains an input-adaptive generator to synthesize targeted perturbation and achieves state-of-the-art transferability. However, a dedicated generator must be learned for every (\textit{source model, target class}) pair in TTP. Such a limitation is partially addressed by training a conditional generator \cite{mi:14} to target multi-class simultaneously (C-GSP, LFAA) \cite{y:22, wanga:23}. However, the number of targeted labels a single generator can cover is limited due to its limited representative capacity. As a consequence, when the number of targeted classes is enormous, e.g., ImageNet, the required training time and storage are still prohibitive.

\begin{figure*}[htb]  
	\centering
	\subfigure[]{
		\includegraphics[width=3.2cm]{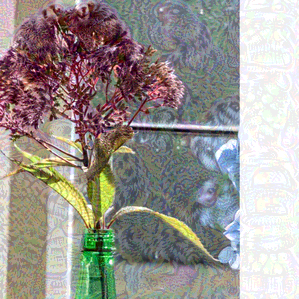}
	}
	\subfigure[]{
		\includegraphics[width=3.2cm]{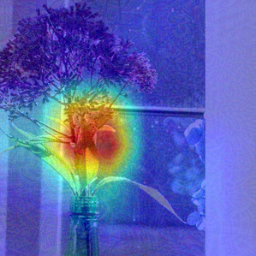}
	}
	\subfigure[]{
		\includegraphics[width=3.2cm]{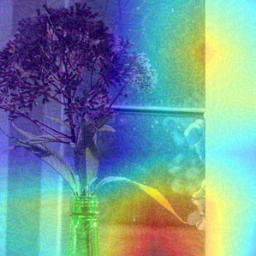}
	}
	\subfigure[]{
		\includegraphics[width=3.2cm]{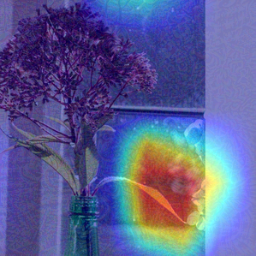}
	}
	\subfigure[]{
		\includegraphics[width=3.2cm]{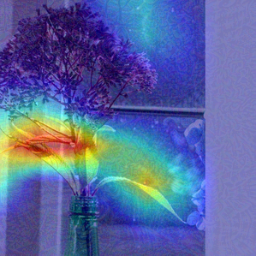}
	}
	\subfigure[]{
		\includegraphics[width=3.2cm]{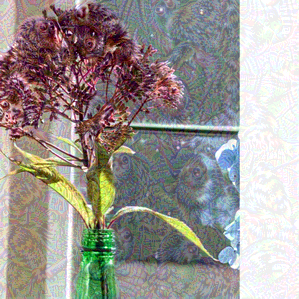}
	}
	\subfigure[]{
		\includegraphics[width=3.2cm]{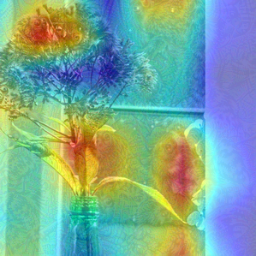}
	}
	\subfigure[]{
		\includegraphics[width=3.2cm]{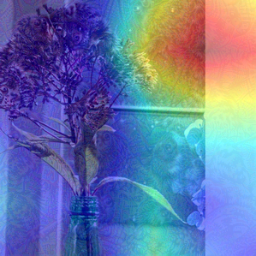}
	}
	\subfigure[]{
		\includegraphics[width=3.2cm]{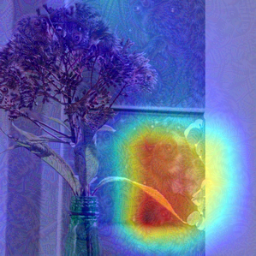}
	}
	\subfigure[]{
		\includegraphics[width=3.2cm]{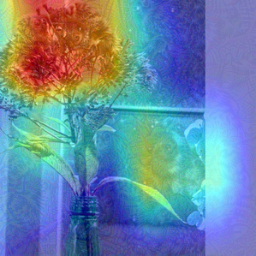}
	}
	\caption{Attentional maps of the target label (`\textit{marmoset}') on different models. The top row depicts the results of the vanilla CE attack, and the bottom that of the proposed CE+\textit{everywhere} attack. (a, f) Crafted AEs, (b, g) VGG16 (surrogate), (c, h) Inceptionv3 (Inc-v3) \cite{sz:16}, (d, i) Res50, (e, j) Dense121.}
\end{figure*}
On the other hand, \textbf{simple-gradient methods} only iteratively optimize a victim image and thus have received more attention. Po+Trip attack \cite{lia:20} replaces traditional cross-entropy (CE) loss with the Poincare distance loss to address the decreasing gradient problem and introduces a triplet loss to push the attacked image away from $y_o$. Logit attack \cite{zh:21} uses the Logit loss in the attack and reports better transferability than the CE loss.
\begin{equation} 
	L_{Logit}=-l_t(\textit{\textbf{I}}')
\end{equation}
\noindent where $l_t(\cdot)$ denotes the logit output with respect to $y_t$. Moreover, Zhao, Liu and Larson (2021) point out that targeted attacks need significantly more iterations to converge than untargeted ones do. Similarly, Weng et al. (2023) (Margin) point out that the vanishing of the logit margin between the targeted and untargeted classes limits targeted transferability. Thus, they downscale the logits with a temperature factor to address the saturation issue and achieve improved transferability. The object-based diverse input method (ODI) \cite{b:22} proposes diversifying the input image in a 3D object manner to avoid overfitting the source model and achieve improved targeted transferability. The high-confidence label suppressing method (SupHigh) \cite{ze:23} argues that not only the original label $y_o$, but other high-confidence labels should also be suppressed for better transferability. Such an idea can be realized by updating AEs according to the following direction:
\begin{equation} 
    \nabla(l_t(\textbf{\textit{I}}')-\beta_1l_o(\textit{\textbf{I}}'))-\beta_2\nabla(\Sigma_{i=0}^{N_h}l_{high-conf,i}(\textit{\textbf{I}}'))\perp
\end{equation}
\noindent where $\perp$ denotes retaining only the component perpendicular to the first item. Here, the first term is used to enhance the confidence of $y_t$ and suppress $y_o$ simultaneously, the second term suppresses other high-confidence labels. Based on the observation that highly universal adversarial perturbations tend to be more transferable, the self-universality method (SU) \cite{wei:23} introduces a feature similarity loss to encourage the adversarial perturbation to be self-universal. The clean feature mixup method (CFM) \cite{b:23} borrowed the idea from Admix \cite{wangb:21} to intentionally introduce competitor noises during optimization, which is achieved by mixing up features from other images in the same batch. Strictly speaking, CFM does not belong to simple-gradient methods since additional images are involved in the optimization. Nevertheless, it exhibits outstanding attack ability according to our experiments.

\section{The Proposed Method}
This section revisits a common cause for targeted transfer failure and details the proposed everywhere attack, which can effectively alleviate the attention mismatch issue.
\begin{figure*}[htb]
	\centering
	\includegraphics[width=0.9\textwidth]{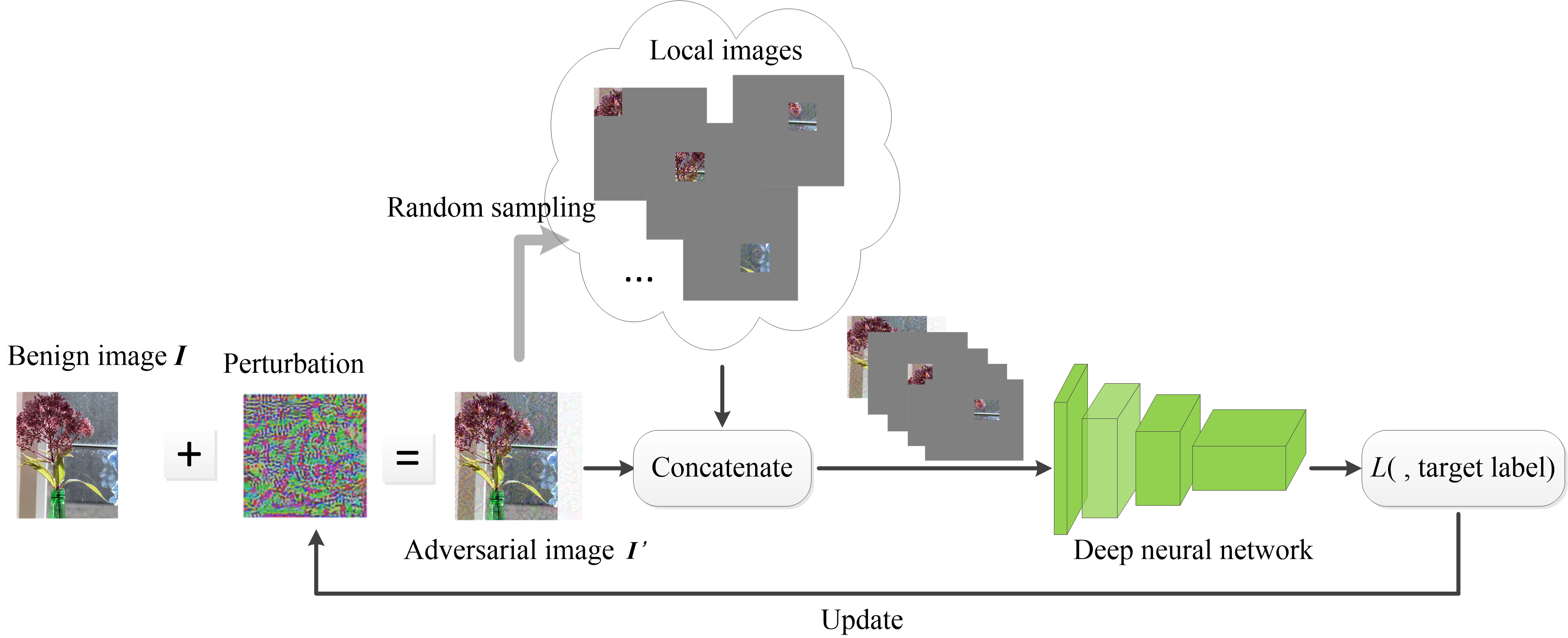} 
	\caption{Overview of the proposed \textit{everywhere} attack.}
	\label{fig3}
\end{figure*}
\subsection{Motivation}
In a targeted attack, the adversary attempts to plant a quasi-imperceptible target object (or objects) into a clean image. Due to the attentional mechanism of DNNs, such a planting often focuses on specific image regions. To achieve transferable attacks across victim models, one may expect victim models to center on regions similar to the surrogate model in identifying the target object (or objects). In fact, this assumption is difficult to satisfy in a targeted attack. To illustrate this dilemma, we examine the attentional maps of an AE on different models. The attentional maps are computed with GradCAM \cite{se:17}. The AE shown in Figure 2(a) is crafted with the vanilla CE attack, the surrogate model is VGG16bn (VGG16) \cite{si:15}, and the target label is `\textit{marmoset}'. As can be observed from Figure 2(b), the attack focuses on the lower area of the flower crown. One can imagine that the adversary has planted a `\textit{marmoset}' in this region. However, victim models pay attention to strikingly different regions in recognizing a `\textit{marmoset}'. For example, ResNet50 (Res50) \cite{h:16} tries to find a `\textit{marmoset}' from the lower right area of the image (Figure 2(d)). As a result, such a transfer attack fails on all three victim models.

One possible way to address the abovementioned challenge is to draw the victim model's attention to the attacked region. However, this is not easy because the adversary in the transfer attack setting cannot access the victim model. Another solution is to craft \textit{a target} in the victim model's attentional region. Since the victim model's attention is unknown in advance, an intuitive strategy is to craft \textit{a bunch of targets} in every region that the victim model may pay attention to. Such a conceptually simple idea motivates the proposed \textit{everywhere} attack.

\subsection{Everywhere Attack}
Figure 3 gives an overview of the proposed \textit{everywhere} attack. To synthesize targets in multiple regions of the image, we split a victim image into $M\times M$ non-overlap blocks. Then, we randomly sample \textit{N} blocks from the image. For each sampled block, we pad the remaining area with the mean value of the dataset (which will be normalized to zero) and get a `local' image. Concatenating these `local' images with the global image delivers \textit{N}+1 images to attack. Finally, we simultaneously mount a targeted attack on these \textit{N}+1 images toward the same target (e.g., `\textit{marmoset}'). In this manner, we expect every block of the obtained AE independently possesses attack capability. The parameter N can be used to balance the attack power and the computational efficiency. Note that the \textit{everywhere} attack degenerates to a baseline attack when $N=0$. Algorithm 1 summarizes the procedure of integrating the proposed \textit{everywhere} scheme with the CE attack, where DI, TI, and MI are conventional transferability-enhanced methods.

The bottom row of Figure 2 shows an AE crafted with the proposed \textit{everywhere} scheme and its attentional maps on different models. The attentional map computed on the surrogate model (Figure 2(g)) presents multiple focal areas. Conceptually, this is similar to the adversary implanting multiple \textit{marmosets} in the image. One of the \textit{marmosets} (the one at the bottom right) is located in the region of interest of Res50 (Figure 2(i)), and another one (the one at the top left) in the region of interest of DenseNet121 (Den121) \cite{h:17} (Figure 2(j)). As a result, our attack successfully transfers to these two victim models.

To conclude this section, we conduct a quantitative experiment on the ImageNet-compatible dataset\footnote{https://github.com/cleverhans-lab/cleverhans/tree/master/ cleverhans$\_$v3.1.0/examples/nips17$\_$adversarial$\_$competition}. We introduce a coverage metric \textit{C} to represent the extent of a victim model'attention ($Att_v$) being covered by that of the surrogate ($Att_s$).
\begin{equation} 
	C=\frac{|Att_v \bigcap Att_s|}{|Att_v|}
\end{equation}
where $Att$ is the normalized ([0, 1]) and binarized (threshold=2/3) attention map. Table 1 reports the averaged coverage metric over 200 images. Obviously, with the \textit{everywhere} scheme, the victim model's attention is more likely to overlap with the surrogate's, i.e., the attention mismatch issue has, in essence, been addressed.

\begin{table}[htb]
	\centering
	\begin{tabular}{c|c|c|c}
		\hline
		& Res50 &Den121 &Inc-v3  \\
		\hline
		CE                     &0.378 &0.383 &0.251 \\
		CE+\textit{everywhere} &0.645 &0.638 &0.504 \\
		\hline	
	\end{tabular}
	\caption{Averaged coverage metric of different victims. Surrogate: VGG16.}
	\label{table1}
\end{table}

\begin{algorithm}[tb]
	\caption{\textit{Everywhere} + CE attack}
	\label{alg:algorithm}
	\textbf{Input}: A benign image \textit{\textbf{I}}; target label $y_t$; a surrogate model \textit{f} with loss function \textit{J}\\
	\textbf{Parameter}: number of partitions for each dimension \textit{M}, samples \textit{N}, iterations \textit{T}\\
	\textbf{Output}: Adversarial Image \textit{\textbf{I}}'
	\begin{algorithmic}[1] 
		\STATE Initialize $\delta_0$ and $g_0$
		\FOR{t=0 to \textit{T}-1}
		\STATE DI: $I_t'=DI(I+\delta_t)$.
		\STATE Split $I_t'$ into $M\times M$ non-overlap blocks.
		\STATE Randomly sample \textit{N} blocks and obtain local images $\textit{\textbf{L}}_0', \textbf{\textit{L}}_1', \cdot \cdot \cdot, \textbf{\textit{L}}_{N-1}'$ by padding.
		\STATE Concatenate: $I_t'=[I_t'$, $\textbf{\textit{L}}_0'$, $\textbf{\textit{L}}_1', \cdot\cdot\cdot, \textbf{\textit{L}}_{N-1}']$.
		\STATE Input $I_t'$ to \textit{f} and obtain gradient $g_{t+1}=\nabla_\delta J(I_t', y_t)$
		\STATE TI and MI: $g_{t+1}=g_t+TI(g_{t+1})$
		\STATE Update and clip $\delta_{t+1}$ 
		\ENDFOR
		\STATE \textbf{return} \textit{\textbf{I}}'=\textit{\textbf{I}}+$\delta_T$
	\end{algorithmic}
\end{algorithm}

\section{Experimental Results}
In this section, we show the efficiency of the proposed \textit{everywhere} attack scheme by integrating it into six iterative attacks: CE, Logit \cite{zh:21}, Margin \cite{Weng:23}, SupHigh \cite{ze:23}, SU \cite{wei:23}, CFM \cite{b:23} on various transfer scenarios. Since more recent targeted attacks have dominated the Po+Trip attack \cite{lia:20}, we omit its results for brevity. All the iterative schemes start with the TMDI attack. Then, we contrast \textit{everywhere} attack with two generative attacks: TTP \cite{n:21} and C-GSP \cite{y:22}. Next, the proposed method is used for crafting Data-free Targeted Universal Adversarial Perturbation (DTUAP) \cite{mo:17, zh:21}, from which our philosophy can be further illustrated. Finally, the crafted AEs are further evaluated using a real-world image recognition system: Google Cloud Vision. The supplementary material provides the ablation study on our key hyper-parameters.

\begin{table*}[htb]
	\centering
	\small
	\begin{tabular}{l|l|l|l|l|l|l|l|l|l|l}
		\hline
		& \multicolumn{5}{c}{Source Model: Res50} & \multicolumn{5}{c}{Source Model: Dense121} \\ 
		\hline
		Attack &$\rightarrow$Inc-v3 &$\rightarrow$Den121 &$\rightarrow$VGG16 &$\rightarrow$Swin &AVG 
		&$\rightarrow$Inc-v3 &$\rightarrow$Res50 &$\rightarrow$VGG16 &$\rightarrow$Swin &AVG \\
		\hline
		CE & 3.9/\textbf{14.1} & 44.9/\textbf{62.3} & 30.5/\textbf{52.2} & 5.2/\textbf{19.0} & 21.1/\textbf{36.8} & 2.8/\textbf{10.3} & 19.0/\textbf{41.7} & 11.3/\textbf{50.6} & 1.8/\textbf{19.2} & 8.7/\textbf{30.5} \\
		Logit & 9.1/\textbf{22.3} & 70.0/\textbf{78.5} & 61.9/\textbf{69.3} & 13.4/\textbf{28.8} & 38.6/\textbf{49.7} & 7.4/\textbf{17.6}&
		42.6/\textbf{58.5} &36.3/\textbf{54.2} &10.5/\textbf{23.8} &24.2/\textbf{38.5} \\
		Margin &10.9/\textbf{21.7} &70.8/\textbf{80.8} &61.2/\textbf{69.4} &16.5/\textbf{33.1} &39.9/\textbf{51.3} &7.6/\textbf{19.8} &
		44.7/\textbf{58.9} &33.4/\textbf{56.4} &11.7/\textbf{24.6} &24.4/\textbf{39.9} \\
		SupHigh &9.9/\textbf{17.8} &74.2/\textbf{82.7} &62.5/\textbf{78.2} &17.1/\textbf{37.3} &40.9/\textbf{54.0} &8.7/\textbf{12.9} &
		47.4/\textbf{64.3} &40.5/\textbf{64.1} &9.3/\textbf{23.6} &26.6/\textbf{41.2} \\
		SU &11.1/\textbf{21.9} &72.5/\textbf{79.2} &63.9/\textbf{67.4} &21.3/\textbf{34.2} &42.2/\textbf{50.7} &10.0/\textbf{17.2} &49.2/\textbf{63.4}
		&42.3/\textbf{55.5} &13.5/\textbf{23.1} &28.8/\textbf{39.8} \\
		CFM &41.4/\textbf{55.3} &83.3/\textbf{87.7} &77.2/\textbf{81.9} &41.5/\textbf{54.2} &60.9/\textbf{69.8} &35.2/\textbf{43.6} &77.3/\textbf{84.8}
		&66.6/\textbf{73.9} &27.1/\textbf{43.4} &51.6/\textbf{61.4} \\
		\hline
		& \multicolumn{5}{c}{Source Model: VGG16} & \multicolumn{5}{c}{Source Model: Inc-v3} \\ 
		\hline
		Attack &$\rightarrow$Inc-v3 &$\rightarrow$Res50 &$\rightarrow$Den121 &$\rightarrow$Swin &AVG &$\rightarrow$Res50 &$\rightarrow$Den121 &$\rightarrow$VGG16 &$\rightarrow$Swin &AVG \\
		\hline
		CE &0.0/\textbf{1.8} &0.3/\textbf{16.4} &0.5/\textbf{15.1} &0.1/\textbf{7.6}
		&0.2/\textbf{10.2} &1.8/\textbf{6.1} &2.5/\textbf{9.6} &1.5/\textbf{7.3} &0.2/\textbf{0.9} &1.5/\textbf{6.0} \\
		Logit
		&0.8/\textbf{3.4} &10.6/\textbf{21.8} &12.8/\textbf{22.3} &6.5/\textbf{13.1} &7.7/\textbf{15.2} &2.4/\textbf{6.8} &3.6/\textbf{14.3} &2.2/\textbf{8.9} &0.2/\textbf{3.4} &2.1/\textbf{8.4} \\
		Margin
		&0.7/\textbf{3.2} &7.9/\textbf{21.1} &12.3/\textbf{18.5} &6.4/\textbf{10.9}
		&6.8/\textbf{13.4} &2.1/\textbf{8.4} &3.2/\textbf{14.6} &1.9/\textbf{9.3}
		&0.9/\textbf{3.0} &2.0/\textbf{8.8} \\
		SupHigh
		&1.1/\textbf{2.6} &11.2/\textbf{18.0} &13.6/\textbf{22.3} &7.0/\textbf{13.7} &8.2/\textbf{14.2} &2.3/\textbf{7.0} &4.5/\textbf{11.5} &2.2/\textbf{9.2}
		&0.3/\textbf{2.3} &2.3/\textbf{7.5} \\
		SU
		&0.9/\textbf{2.2} &13.7/\textbf{25.2} &15.7/\textbf{24.6} &8.1/\textbf{11.8}
		&9.6/\textbf{16.0} &3.0/\textbf{7.4} &4.6/\textbf{11.9} &3.5/\textbf{8.6}
		&0.9/\textbf{2.8} &3.0/\textbf{7.8} \\
		CFM
		&3.8/\textbf{9.3} &26.1/\textbf{34.7} &28.3/\textbf{39.5} &12.4/\textbf{20.8}
		&17.7/\textbf{26.1} &12.3/\textbf{29.8} &20.9/\textbf{40.3} &13.4/\textbf{25.6}
		&4.0/\textbf{11.4} &12.7/\textbf{26.8}  \\
		\hline
	\end{tabular}
\caption{Targeted transfer success rate (\%) without/with the proposed \textit{everywhere} scheme, in the random-target scenario. The AVG column is averaged over victims. Best results are in \textbf{bold}.}
\label{table2}
\end{table*}
\subsection{Experimental Settings}
\textbf{Dataset}. Following recent work on targeted attacks, our experiments are conducted on the ImageNet-compatible dataset comprised of 1000 images. All these images are with the size of $299\times 299$ pixels and are stored in PNG format.

\noindent\textbf{Networks}. Since transferring across different architectures is more demanding, we choose four pretrained models of diverse architectures: Inc-v3, Res50, Den121, and VGG16 as the surrogates. These surrogates and a transformer-based model, Swin \cite{liu:21}, evaluate AEs' transferability.

\noindent\textbf{Parameters}. For all attacks, the perturbations are restricted by $L_\infty$ norm with $\epsilon=16$ (The results under lower budgets are provided in the supplementary material), and the step size is set to 2. The total iteration number \textit{T} is set to 200 to balance speed and convergence. The number of partitions for each dimension \textit{M} is set to 4, and the number of samples \textit{N} is set to 9.

\subsection{Normal surrogates}
Table 2 reports the targeted transferability (random-target) across different models. The proposed\textit{ everywhere} scheme boosts all the baseline attacks by a clear margin. Taking the popular Logit attack as a baseline, the average success rate has been improved by 28.8\% (49.7\% vs. 38.6\%) $\sim$ 300\% (8.4\% vs. 2.1\%). Further analysis can provide more insights into the proposed method. First, the weaker the baseline, the more significant the improvement. Hence, the upturn is particularly salient for the CE attack. For example, when VGG16 was the surrogate model, the average success rate of the CE attack increases from 0.2\% to 10.2\%. Second, the more challenging the transfer scenario is, the more significant the improvement brought by the proposed \textit{everywhere} scheme. For example, the introduced improvement in the `Res50$\rightarrow$Swin' scenario is much more salient than that in the `Res50$\rightarrow$Dense121', which makes the proposed method even more promising with the popularity of transformer-based networks.

\begin{table*}[htb]
	\centering
	\small
	\begin{tabular}{l|l|l|l|l|l|l|l|l|l|l}
		\hline
		& \multicolumn{5}{c}{Source Model: Res50} & \multicolumn{5}{c}{Source Model: Dense121} \\ 
		\hline
		Attack &$\rightarrow$Inc-v3 &$\rightarrow$Den121 &$\rightarrow$VGG16 &$\rightarrow$Swin &AVG &$\rightarrow$Inc-v3 &$\rightarrow$Res50 &$\rightarrow$VGG16 &$\rightarrow$Swin &AVG \\
		\hline
		CE
		&1.3/\textbf{8.8} &25.8/\textbf{52.1} &15.0/\textbf{42.3}
		&3.2/\textbf{19.9} &11.3/\textbf{30.8} &1.2/\textbf{6.1}
		&6.5/\textbf{32.7} &3.6/\textbf{36.2} &0.6/\textbf{12.6} &3.0/\textbf{21.9} \\

		Logit
		&3.6/\textbf{9.2} &51.6/\textbf{64.7} &38.6/\textbf{47.1}
		&9.2/\textbf{23.2} &25.8/\textbf{36.1} &3.5/\textbf{10.2}
		&22.7/\textbf{46.1} &18.3/\textbf{34.9} &4.7/\textbf{14.3} &12.3/\textbf{26.4} \\

		Margin 
		&4.1/\textbf{12.1} &52.3/\textbf{65.8} &38.9/\textbf{47.5}
	    &10.2/\textbf{22.4} &26.5/\textbf{37.0} &3.9/\textbf{9.5}
		&24.4/\textbf{44.2} &18.2/\textbf{41.3} &5.1/\textbf{15.6}
		&12.9/\textbf{27.7} \\

		SupHigh 
		&4.0/\textbf{8.8} &53.5/\textbf{68.6} &41.6/\textbf{60.1}
		&8.1/\textbf{24.8} &26.8/\textbf{40.6} &3.8/\textbf{7.2}
		&24.5/\textbf{51.1} &21.2/\textbf{42.5} &5.2/\textbf{15.7}
		&13.7/\textbf{29.1} \\
		SU 
		&5.3/\textbf{11.2} &54.2/\textbf{66.7} &44.1/\textbf{48.6}
		&13.2/\textbf{22.3} &29.2/\textbf{37.2} &4.4/\textbf{11.3}
		&27.4/\textbf{44.7} &24.3/\textbf{39.9} &9.0/\textbf{16.8}
		&16.3/\textbf{28.2} \\
		CFM 
		&28.2/\textbf{37.3} &76.9/\textbf{84.8} &61.8/\textbf{70.1}
		&24.5/\textbf{44.2} &47.9/\textbf{59.1} &27.3/\textbf{36.2}
		&66.1/\textbf{72.6} &51.8/\textbf{61.7} &21.8/\textbf{36.0}
		&41.8/\textbf{51.6} \\
		\hline
		& \multicolumn{5}{c}{Source Model: VGG16} & \multicolumn{5}{c}{Source Model: Inc-v3} \\ 
		\hline
		Attack &$\rightarrow$Inc-v3 &$\rightarrow$Res50 &$\rightarrow$Den121 &$\rightarrow$Swin &AVG &$\rightarrow$Res50 &$\rightarrow$Den121 &$\rightarrow$VGG16 &$\rightarrow$Swin &AVG \\
		\hline
		CE
		&0.0/\textbf{1.1} &0.0/\textbf{3.6} &0.0/\textbf{6.4}
		&0.0/\textbf{4.8} &0.0/\textbf{4.0} &2.4/\textbf{7.8}
		&3.8/\textbf{9.1} &2.3/\textbf{5.7} &0.6/\textbf{2.6}
		&2.3/\textbf{6.3} \\
		Logit
		&0.3/\textbf{0.7} &3.3/\textbf{10.9} &6.8/\textbf{12.1}
		&5.0/\textbf{11.9} &3.9/\textbf{8.9} &3.8/\textbf{10.9}
		&4.5/\textbf{10.6} &3.2/\textbf{8.3} &0.5/\textbf{2.7}
		&3.0/\textbf{8.1} \\
		Margin
		&0.0/\textbf{0.4} &4.4/\textbf{6.8} &6.3/\textbf{9.9}
		&6.4/\textbf{8.1} &4.3/\textbf{6.3} &2.5/\textbf{13.2}
		&4.3/\textbf{13.8} &2.0/\textbf{10.9} &0.2/\textbf{2.6}
		&2.3/\textbf{10.1} \\
		SupHigh
		&0.1\textbf{/0.3} &3.9/\textbf{7.1} &6.8/\textbf{8.7}
		&3.1/\textbf{10.2} &3.5/\textbf{6.6} &3.5/\textbf{10.8}
		&4.9/\textbf{16.7} &3.4/\textbf{11.3} &0.4/\textbf{3.9}
		&3.1/\textbf{10.7} \\
		SU
		&0.3/\textbf{0.7} &5.7/\textbf{9.8} &7.4/\textbf{16.8}
		&4.5/\textbf{11.1} &4.5/\textbf{9.6} &4.3/\textbf{10.2}
		&6.7/\textbf{13.2} &3.9/\textbf{8.1} &0.8/\textbf{2.1}
		&3.9/\textbf{8.4} \\
		CFM
		&2.7/\textbf{4.2} &13.1/\textbf{21.8} &17.3/\textbf{28.5}
		&8.6/\textbf{14.9} &10.4/\textbf{17.4} &16.7/\textbf{35.3}
		&22.2/\textbf{42.1} &10.2/\textbf{25.8} &4.6/\textbf{17.4}
		&13.4/\textbf{30.2} \\
		\hline
	\end{tabular}
\caption{Targeted transfer success rate (\%) without/with the proposed \textit{everywhere} scheme, in the most difficult-target scenario.}
\label{table3}
\end{table*}

As done in previous works \cite{zh:21, ze:23}, we also conduct a worst-case transfer experiment in which the target labels are always the least likely ones. Table 3 compares different attacks: the improvement from the proposed \textit{everywhere} scheme is even more remarkable than the random-target scenario. Taking the Logit attack as the baseline again, the average success rate increases by 39.9\% (36.1\% vs. 25.8\%) when Res50 is the surrogate, and it more than doubles for other surrogates.

\begin{table*}[htb]
	\centering
	\small
	\begin{tabular}{l|l|l|l|l|l|l|l|l|l|l}
		\hline
		& \multicolumn{5}{c}{Source Model: Res18adv} 
		& \multicolumn{5}{c}{Source Model: Res50adv} \\ 
		\hline
		Attack &$\rightarrow$Inc-v3 &$\rightarrow$Den121 &$\rightarrow$VGG16 &$\rightarrow$Swin &AVG &$\rightarrow$Inc-v3 &$\rightarrow$Den121 &$\rightarrow$VGG16 &$\rightarrow$Swin &AVG \\
		\hline
		CE
		&6.7/\textbf{13.2} &29.4/\textbf{44.6} &13.2/\textbf{35.9}
		&2.4/\textbf{16.3} &12.9/\textbf{27.5} &14.4/\textbf{16.9}
		&59.0/\textbf{64.4} &24.8/\textbf{53.1} &6.6/\textbf{28.8}
		&26.2/\textbf{40.8} \\
		
		Logit
		&21.8/\textbf{27.0} &60.3/\textbf{68.2} &46.2/\textbf{50.7}
		&13.1/\textbf{25.7} &35.4/\textbf{42.9} &26.1/\textbf{30.8}
		&78.6/\textbf{83.9} &55.9/\textbf{67.4} &23.2/\textbf{41.6}
		&46.0/\textbf{55.9} \\
		
		Margin 
		&20.4/\textbf{22.4} &62.5/\textbf{65.1} &43.6/\textbf{51.2}
		&14.2/\textbf{21.1} &35.2/\textbf{39.9} &26.8/\textbf{29.3}
		&82.3/\textbf{83.6} &55.6/\textbf{67.5} &25.3/\textbf{38.0}
		&47.5/\textbf{54.6} \\
		
		SupHigh 
		&21.0/\textbf{29.4} &68.6/\textbf{75.6} &56.1/\textbf{65.4}
		&20.2/\textbf{32.3} &41.5/\textbf{50.7} &21.4/\textbf{27.5}
		&80.7/\textbf{87.0} &67.8/\textbf{76.9} &29.1/\textbf{48.2}
		&49.7/\textbf{59.9} \\
		
		SU 
		&23.4/\textbf{30.3} &65.8/\textbf{70.3} &45.3/\textbf{51.8}
		&15.9/\textbf{25.9} &37.4/\textbf{49.6} &27.6/\textbf{29.5}
		&79.9/\textbf{81.3} &56.8/\textbf{64.2} &24.5/\textbf{41.4}
		&47.2/\textbf{54.2} \\
		
		CFM 
		&36.1/\textbf{41.2} &75.6/\textbf{80.4} &56.7/\textbf{64.6}
		&27.8/\textbf{38.1} &49.1/\textbf{56.2} &51.8/\textbf{58.3}
		&85.6/\textbf{86.9} &74.7/\textbf{79.1} &47.5/\textbf{60.2}
		&64.9/\textbf{71.2} \\
		\hline	
	\end{tabular}
\caption{Targeted transfer success rate (\%) without/with the \textit{everywhere} scheme. The AEs are crafted against robust models.}
\label{table4}
\end{table*}

\subsection{Robust surrogates}
Leveraging a slightly robust (adversarially trained) surrogate is accepted as an efficient way to craft transferable targeted AEs \cite{sp:21}. We are interested in how the proposed \textit{everywhere} scheme can improve the baselines when robust models are used as surrogates. Specifically, AEs are crafted with adversarially trained models Res18adv and Res50adv and transferred to the same victims used in the last section except Res50. Both models are trained with AEs under $L_2=0.01$ budget. Note that there is no architectural overlap between the source and target models. Table 4 presents the targeted transferability in this case. Even though AEs crafted by robust models have shown significantly stronger transferability than those crafted with normal surrogates, the proposed \textit{everywhere} scheme is still helpful, especially when the transformer-based model Swin is the victim. Taking the Logit attack for example, the targeted success rate is doubled in the `Res18adv$\rightarrow$Swin' scenario (25.7\% vs. 13.1\%) and improved by more than a half in the `Res50adv$\rightarrow$Swin' scenario (41.6\% vs. 23.2\%).

\begin{table}[htb]
	\centering
	\small
	\begin{tabular}{l|l|l|l|l|l}
		\hline
		Attack &Inc-v3 &Den121 &VGG16 &Swin &AVG  \\
		\hline
		CE
		&15.1 &63.3 &58.8 &23.6 &40.2 \\
		Logit
		&22.8 &83.1 &74.0 &35.8 &53.9 \\
		Margin 
		&23.4 &83.3 &75.4 &38.3 &55.1 \\
		SupHigh 
		&19.0 &87.5 &85.5 &39.4 &57.9 \\
		SU 
		&24.7 &83.1 &74.0 &36.7 &54.6 \\
		CFM 
		&\textbf{59.0} &\textbf{93.7} &\textbf{90.0} &\textbf{59.3} &\textbf{75.5} \\
		\hline
		TTP
		&39.8 &79.5 &75.4 &44.6 &59.8 \\
		C-GSP
		&30.1 &67.5 &57.0 &34.4 &47.3 \\
		\hline	
	\end{tabular}
\caption{Iterative attacks vs. generative attacks, The transfer success rates (\%) are averaged over 10 target classes. The upper part of the table presents the results of six iterative attacks, while the lower part shows the results of two generative attacks. Iterative attacks are integrated with the proposed \textit{everywhere} scheme, and the source model is Res50.}
\label{table5}
\end{table}
\subsection{Iterative vs. generative attacks}
Next, we compare the proposed \textit{everywhere} attack with the state-of-the-art generative attacks, TTP and C-GSP. As mentioned before, TTP entails training a generator for each target label and each source model. That means $4\times 1000$ generators are required to perform the random or most difficult-target attack, which is computationally prohibitive. Alternatively, we follow the `\textit{10-Targets (all source)}' setting of \cite{n:21} and use ten author-released generators (Res50 being the discriminator during training) to generate AEs. For C-GSP, we train a 10-target conditional generator with Res50 being the discriminator on the ImageNet `\textit{train}' dataset \cite{r:15}.

As shown in Table 5, between two generative methods, the attack ability of the multi-class generator is inevitably inferior to that of the single-class generator. Nevertheless, with the proposed \textit{everywhere} scheme, iterative attacks may yield comparable or even better (CFM + \textit{everywhere}) transferability than generative methods. Such results demonstrate the potential of iterative attacks in the face of generative ones. However, we must admit the intrinsic advantage of the generative attacks: Once the generators are trained, they can craft AEs with much higher computational efficiency than iterative attacks.

\begin{table}[htb]
	\centering
	\begin{tabular}{l|l|l|l|l}
		\hline
		&Res50 &Den121 &VGG16 &Inc-v3 \\
		\hline
		CE
		&8.1/\textbf{19.4} &8.0/\textbf{28.3} &19.2/\textbf{61.5} &1.9/\textbf{4.8} \\
		Logit
		&20.7/\textbf{25.1} &17.5/\textbf{27.3} &64.9/\textbf{70.5} &3.6/\textbf{5.0}\\
		\hline	
	\end{tabular}
	\caption{Success rates (\%) of the data-free UAPs with $\epsilon=16$, without/with the proposed \textit{everywhere} scheme.}
	\label{table6}
\end{table}
\begin{figure*}[htb]  
	\centering
	\subfigure[]{
		\includegraphics[width=3.2cm]{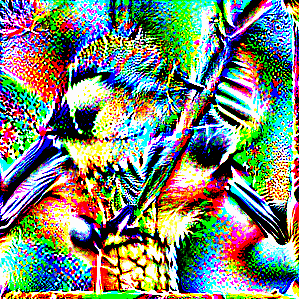}
	}
	\subfigure[]{
		\includegraphics[width=3.2cm]{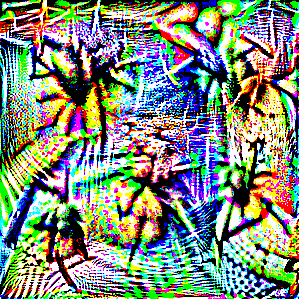}
	}
	\subfigure[]{
		\includegraphics[width=3.2cm]{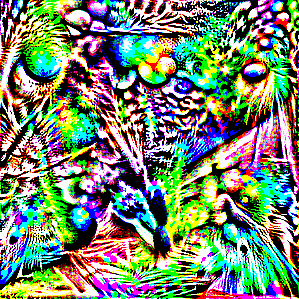}
	}
	\subfigure[]{
		\includegraphics[width=3.2cm]{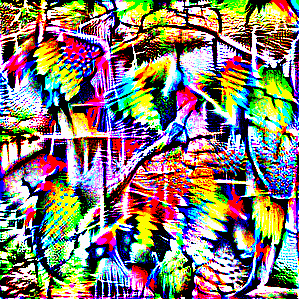}
	}
	\subfigure[]{
		\includegraphics[width=3.2cm]{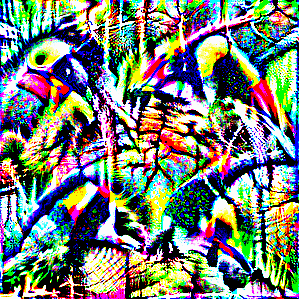}
	}
	\subfigure[]{
		\includegraphics[width=3.2cm]{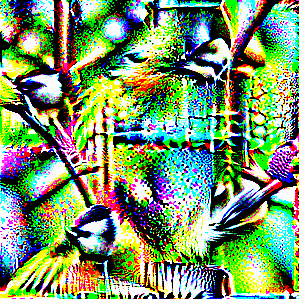}
	}
	\subfigure[]{
		\includegraphics[width=3.2cm]{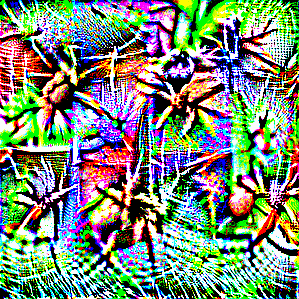}
	}
	\subfigure[]{
		\includegraphics[width=3.2cm]{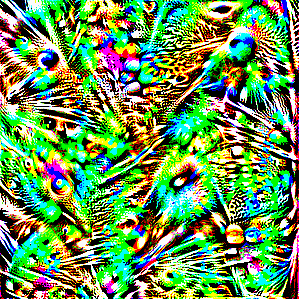}
	}
	\subfigure[]{
		\includegraphics[width=3.2cm]{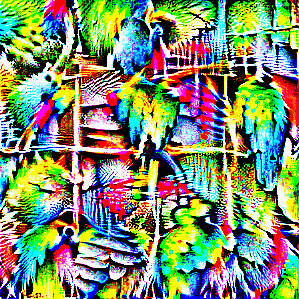}
	}
	\subfigure[]{
		\includegraphics[width=3.2cm]{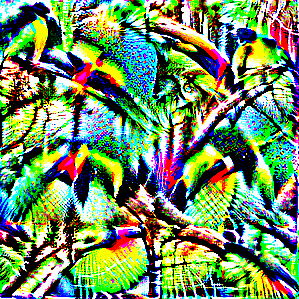}
	}
	\caption{Data-free UAPs of different target classes using Logit (top) and Logit+\textit{everywhere} (bottom). (a, f) `chickadee', (b, g) `wolf spider', (c, h) `peacock', (d, i) `macaw', (e, j) `toucan'. The UAPs have been scaled to [0, 1] for better visualization.}
\end{figure*}
\subsection{Data-free Targeted UAP}
DTUAP is optimized from a random image and can drive multiple clean images into a given class $y_t$. Due to its data-free nature, DTUAP is a powerful tool for uncovering the intrinsic features of the model of interest. Following \cite{zh:21}, we use a mean image (all entrances of which equal 0.5) as the starting point and mount a targeted attack to obtain a DTUAP with CE and Logit attacks ($\epsilon$=16). Then, the obtained DTUAP is applied to all 1000 images in our dataset. Table 6 reports the success rates averaged over 100 classes ($y_t=0:99$). It is observed that the proposed \textit{everywhere} scheme yields more transferable UAPs across input images compared with baselines. For example, with the Logit+\textit{everywhere} scheme, the DTUAPs crafted on the VGG16 model can successfully drive seventy percent of the images into a specified class.

To provide a more intuitive explanation of the proposed everywhere scheme, we depict several DTUAP samples in Figure 4. Since features learned by the robust models are more semantically aligned, here we use Res50adv to craft DTUAPs. Compared to the baseline attack, the \textit{everywhere} attack tends to plant more target objects with smaller sizes into the obtained UAP. Such a distinction is apparent in the case of `\textit{chickadee}'. Only one big \textit{chickadee} can be observed in the DTUAP crafted by the vanilla Logit attack (Figure 4(a)). In contrast, at least four baby \textit{chickadees} can be found in the DTUAP crafted by Logit+\textit{everywhere} attack (Figure 4(f)).

\begin{table}[htb]
	\centering
	\begin{tabular}{c|c|c|c}
		\hline
		Logit &Logit+\textit{everywhere} &CFM &CFM+\textit{everywhere} \\
		\hline
		6 &11 &26 &47 \\
		\hline	
	\end{tabular}
	\caption{Success rates (\%) of different attacks on Google Cloud Vision. Surrogate: Res50adv.}
	\label{table7}
\end{table}
\subsection{Fooling Google Cloud Vision}
Finally, we evaluate the crafted AEs on the Google Cloud Vision API. Specifically, targeted AEs are generated with Res50adv, and the API returns a list of semantic labels for each probe image. As \cite{zh:21}, the attack is deemed a success once the target appears in the returned list. Note that we regard semantically similar classes as the same since the semantic label set of the API does not precisely match the ImageNet classes.

Table 7 reports the targeted success rate averaged over 100 images. Google Cloud Vision API is much more difficult to transfer than previously studied models. Nevertheless, the proposed \textit{everywhere} scheme effectively improves the transferability of baselines. Due to page limitations, we provide the sample images and the API outputs in the supplementary material.

\section{Conclusion}
The discriminative regions of a target class on victim models are dramatically different from that on the surrogate, which severely constrains the targeted transferability of AEs. To address this challenge, we propose the \textit{everywhere} attack, which optimizes an army of target objects in every local image region that victim models may pay attention to and thus reduces the transfer failures caused by attention mismatch. Extensive experiments demonstrate that the proposed method can universally boost the transferability of existing targeted attacks. It is our hope that the idea of increasing the target quantity opens a new door to boosting targeted transferability for the community.
 
\section{Acknowledgments}
This work was supported by the Opening Project of Guangdong Province
Key Laboratory of Information Security Technology (no. 2023B1212060026) and the Start-up Scientific Research Project for Introducing Talents of Beijing Normal University at Zhuhai (no. 312200502504).

Salute to Mr. Nai'an, the author of the novel \textit{Journey to the West}, for his 520th anniversary of birth.

\bibliography{aaai25_everywhere}

\newpage
\appendix
\section{Supplementary Material}
The supplementary document consists of four parts of content: A) Ablation studies on \textit{M} and \textit{N}; B) Attacking transformer-based models; C) A theoretical analysis of the \textit{everywhere} scheme; and D) Adversarial examples (AE) on Google Cloud Vision.

\section{Ablation study}
1) \textbf{Influence of the number of samples} \textit{N}. \textit{N} indicates how many local blocks are sampled (out of $M^2$) to attack in each iteration. A small \textit{N} may cause an underattack in each iteration and need more iterations to converge, whereas a large \textit{N} consumes more memory. Baseline+\textit{everywhere} attack reduces to the baseline attack when $N=0$.

We study the influence of \textit{N} of the proposed Logit+\textit{everywhere}
attack in the random-target scenario. The reported attack success rates are averaged over four victims, e.g., Res50, Dense121, VGG16, and Swin when the surrogate is Inc-v3. The number of partitions \textit{M} for each dimension is
fixed as 4; thus, \textit{N} varies from 0 to 16. As can be observed from Figure 1(a), the average success rates grow steadily at the beginning and tend to saturate after $N\geq10$. In our study, we set $N=9$ to balance memory consumption and attack ability.
\begin{figure*}[htb]  
	\centering
	\subfigure[]{
		\includegraphics[width=8cm]{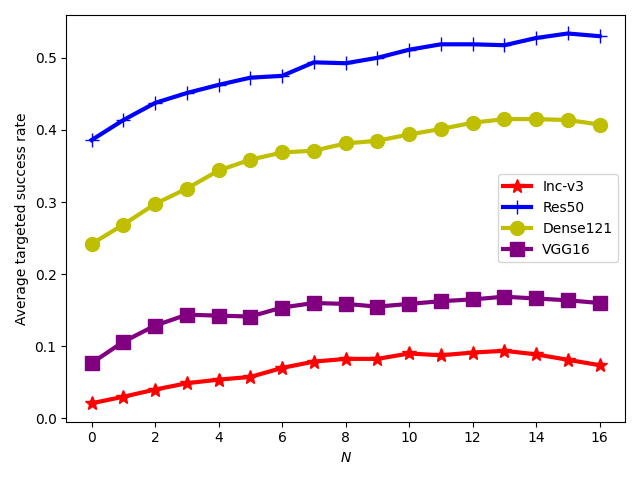}
	}
	\subfigure[]{
		\includegraphics[width=8cm]{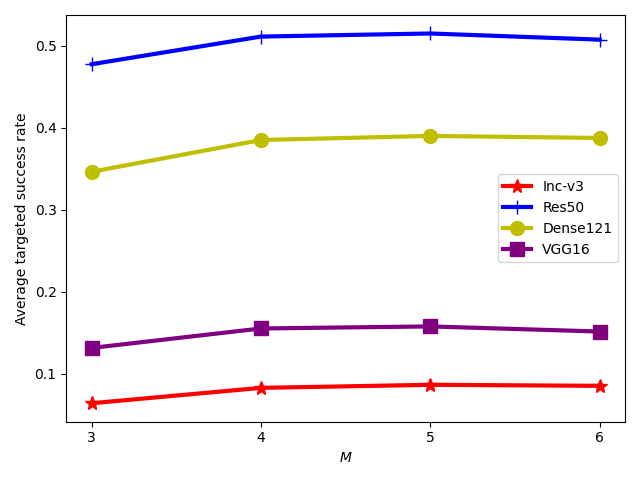}
	}
	\begin{center}
		Figure 1: Ablation study on our newly introduced hyperparameters. Effect of the number of samples \textit{N} (a), and the number of partitions \textit{M} (b) on AEs' transferability. The baseline attack is Logit, and each line corresponds to a different surrogate.
	\end{center}
\end{figure*}

2) \textbf{Influence of the number of partitions \textit{M} for each dimension}. Next, we fix $N=9$ and let \textit{M} vary from 3 to 6 (Note $M^2 \geq N$). Smaller \textit{M} indicates larger size of the local images (before padding) and $M=1$ means attacking the global image only. On the other hand, larger \textit{M} indicates smaller local images
and more attack iterations may be required to converge. To avoid the study overwhelming, we set the number of iterations $T=200$ in all cases.

Figure 1(b) shows the average success rates of different surrogates as functions of \textit{M}. It can be observed that the attack ability of the proposed method is insensitive to \textit{M}. The only exception is $M=3$, in which the lack of randomness leads to inferior transferability. In our study, we set $M=4$ for all attacks and in all scenarios for simplicity.

\section{Attacking transformers}
Table 1 reports the targeted transferability against three transformer-based models, vit\_b\_16 (Dosovitskiy et al. 2021), pit\_b\_24 (Heo et al. 2021), and visformer (Chen et al. 2021), in the random-target scenario. Compared to the results on CNNs (Table 2 of the paper), the improvement introduced by \textit{everywhere} attack is more remarkable when the victim is a transformer. Taking the Logit attack as a baseline, the average success rate has been improved by 66.7\% (1.0\%
vs. 0.6\%) $\sim$ 175\% (7.7\% vs. 2.8\%). We speculate that this is because the blockwise attack strategy in our method is more consistent with the way the transformer understands the image. 

An interesting observation is that vit\_b\_16 and pit\_b\_24 are much more resilient under attack than visformer, which deserves future study.\\[0.1cm]

\noindent Dosovitskiy, A.; Beyer, L.; Kolesnikov, A.; et al. 2021. An image is worth 16x16 words: Transformers for image recognition at scale. In \textit{ICLR}.

\noindent Heo, B.; Yun, S.; Han, D.; et al. 2021. Rethinking spatial dimensions of vision transformers. In \textit{ICCV}, pp. 11916--11925.

\noindent Chen, Z.; Xie, L.; Niu, J.; et al. 2021. Visformer: The vision-friendly transformer. In \textit{ICCV}, pp. 569--578.

\begin{table*}[htb]
	\centering
	\small
	\begin{tabular}{l|l|l|l|l|l|l|l|l}
		\hline
		& \multicolumn{4}{c}{Source Model: Res50} & \multicolumn{4}{c}{Source Model: Dense121} \\ 
		\hline
		Attack &$\rightarrow$vit\_b\_16 &$\rightarrow$pit\_b\_24 &$\rightarrow$visformer  &AVG 
		&$\rightarrow$vit\_b\_16 &$\rightarrow$pit\_b\_24 &$\rightarrow$visformer &AVG \\
		\hline
		CE 
		&0.6/\textbf{3.7} &2.0/\textbf{3.5} &4.8/\textbf{15.3} &2.5/\textbf{7.5}  &1.2/\textbf{3.1} &1.2/\textbf{2.7} &6.2/\textbf{22.4} &2.9/\textbf{9.4} \\
		Logit 
		&2.7/\textbf{9.2} &6.0/\textbf{13.4} &16.0/\textbf{32.2} &8.2/\textbf{18.3}  
		&2.5/\textbf{6.2} &4.7/\textbf{8.9} &23.5/\textbf{37.4} &10.2/\textbf{17.5} \\
		Margin 
		&4.8/\textbf{6.4} &7.6/\textbf{9.3} &19.5/\textbf{28.4} &10.6/\textbf{14.7} 
		&3.6/\textbf{7.2} &5.2/\textbf{7.4} &20.8/\textbf{31.8} &9.9/\textbf{15.5} \\
		SH 
		&3.7/\textbf{6.6} &7.3/\textbf{18.8} &20.1/\textbf{36.1} &10.4/\textbf{20.5}  
		&2.9/\textbf{7.9} &4.0/\textbf{12.6} &25.2/\textbf{38.9} &10.7/\textbf{19.8}  \\
		SU 
		&5.0/\textbf{5.3} &4.8/\textbf{12.9} &20.0/\textbf{29.6} &9.9/\textbf{15.9} 
		&4.4/\textbf{5.8} &4.4/\textbf{4.9} &23.9/\textbf{29.0} &10.9/\textbf{13.2}  \\
		\hline
		& \multicolumn{4}{c}{Source Model: VGG16} & \multicolumn{4}{c}{Source Model: Inc-v3} \\ 
		\hline
		Attack &$\rightarrow$vit\_b\_16 &$\rightarrow$pit\_b\_24 &$\rightarrow$visformer &AVG &$\rightarrow$vit\_b\_16 &$\rightarrow$pit\_b\_24 &$\rightarrow$visformer &AVG \\
		\hline
		CE 
		&0.0/\textbf{0.6} &0.0/\textbf{0.5} &0.6/\textbf{7.3} &0.2/\textbf{2.8}
	    &0.2/\textbf{0.4} &0.2/\textbf{0.5} &0.7/\textbf{1.3} &0.4/\textbf{0.7} \\
		Logit
		&0.2/\textbf{1.2} &1.4/\textbf{4.4} &6.7/\textbf{17.6} &2.8/\textbf{7.7} &0.3/\textbf{0.8} &0.6/\textbf{0.7} &1.0/\textbf{1.5} &0.6/\textbf{1.0} \\
		Margin
		&0.1/\textbf{0.8} &1.8/\textbf{3.3} &4.2/\textbf{9.2} &2.0/\textbf{4.4}
		&0.4/\textbf{0.6} &0.4/\textbf{1.2} &0.9/\textbf{1.6} &0.6/\textbf{1.1} \\
		SH
		&0.4/\textbf{0.9} &2.0/\textbf{5.9} &9.4/\textbf{15.3} &3.9/\textbf{7.4}  &0.2/\textbf{1.6} &0.7/\textbf{0.8} &0.8/\textbf{2.2} &0.6/\textbf{1.5} \\
		SU
		&0.8/\textbf{1.3} &2.2/\textbf{6.2} &12.7/\textbf{14.2} &5.2/\textbf{7.2}
	    &0.2/\textbf{0.7} &0.2/\textbf{0.9} &1.0/\textbf{1.5} &0.5/\textbf{1.9} \\
		\hline
	\end{tabular}
    \begin{center}
    	Table 1: Targeted transfer success rate (\%) $w. o./w.$ the \textit{everywhere} scheme against transformers, in the random-target
    	scenario. The images are down-sampled to the size of $224\times224$ pixels from the original $299\times299$ pixels.
    \end{center}
\label{tableA1}
\end{table*}

\section{How can \textit{everywhere} improve transferability?}
Besides the experimental evidence of the power of the proposed \textit{everywhere} attack, a theoretical analysis of it is provided
in the following.

1) \textit{Everywhere} attack optimizes an army of targets in different regions of the victim image, which can mitigate potential failures caused by the attention mismatch between surrogate and target models.

2) Traditional methods synthesize image-level target-related features in crafting AEs. To a great extent, their attack ability relies on complicated, large-scale interactions between different image regions, which have been proven to be negatively correlated to adversarial transferability (Wang et al. 2021). In contrast, the proposed \textit{everywhere} attack focuses on local features and ignores those fragile large-scale interactions. Thus, stronger transferability is expected.\\[0.2cm]

\noindent Wang, X.; Ren, J.; Lin, S.; et al. 2021. A unified approach to interpreting and boosting adversarial transferability. In \textit{ICLR}.

\section{Adversarial examples on Google Cloud Vision}
Here, we provide a few examples for the paper's `Fooling Google Cloud Vision' section. The left column of Figure 2 shows AEs crafted with the CFM attack, which only succeeds in the second case (`strawberry'$\rightarrow$`tench'). The results of the proposed CFM+\textit{everywhere} attack are shown in the
right column, where all AEs are predicted as the adversary-desired classes with high confidence by the Google Cloud Vision API. For example, in the first case, the API predicts our crafted image as `American lobster' with a confidence
of 0.89.

\begin{figure*}[htb]  
	\centering
	\subfigure[]{
		\includegraphics[width=8cm]{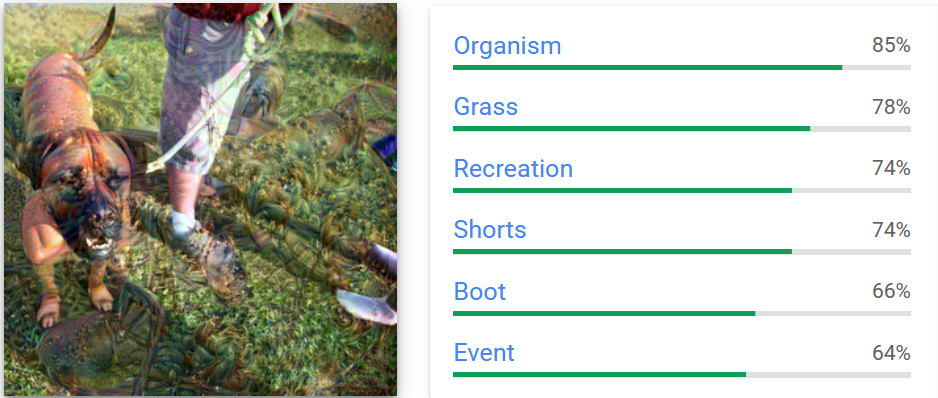}
	}
	\subfigure[]{
		\includegraphics[width=8cm]{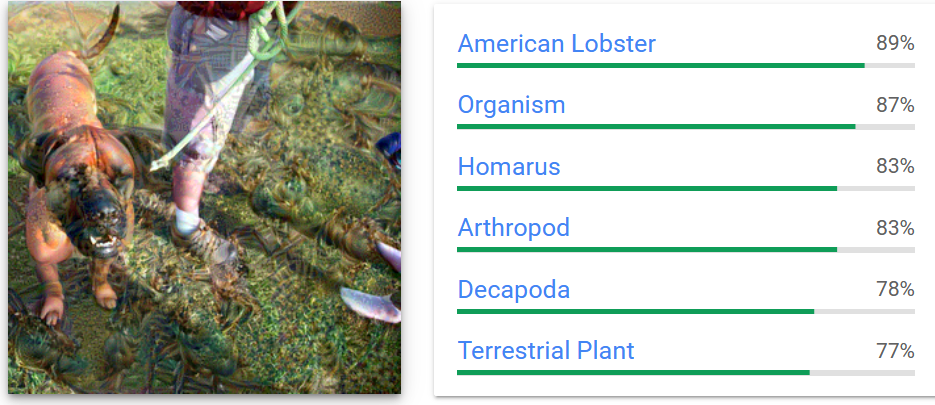}
	}
    \subfigure[]{
    	\includegraphics[width=8cm]{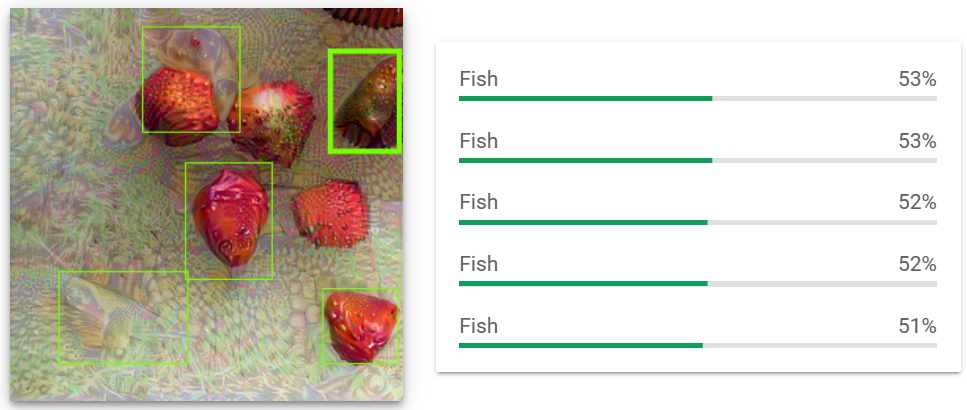}
    }
    \subfigure[]{
    	\includegraphics[width=8cm]{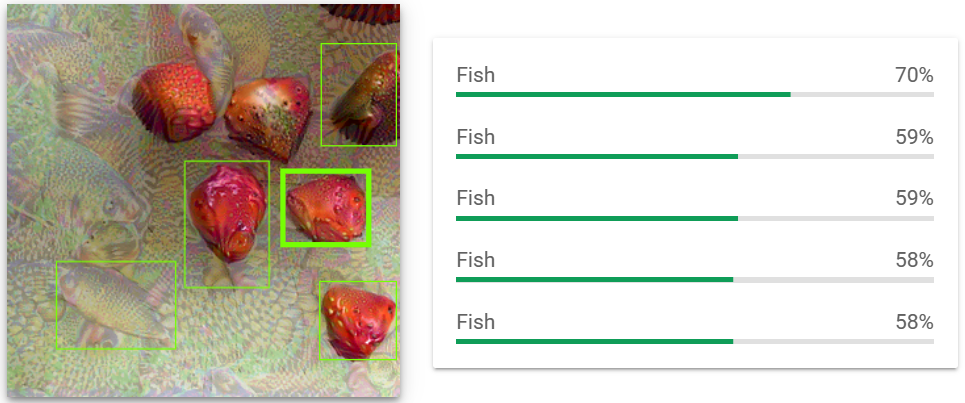}
    }
    \subfigure[]{
	    \includegraphics[width=8cm]{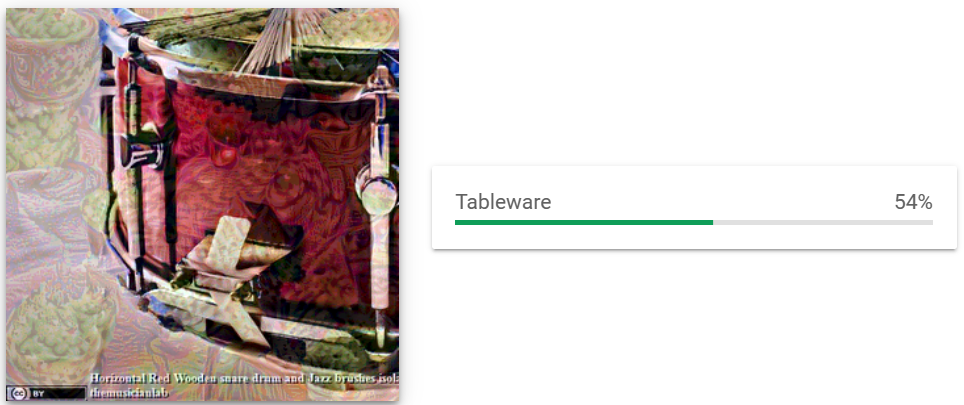}
    }
    \subfigure[]{
	    \includegraphics[width=8cm]{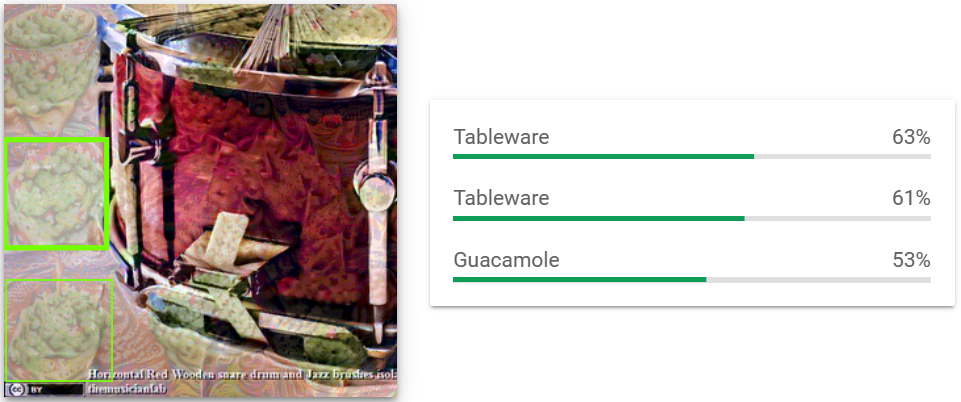}
    }
    \subfigure[]{
    	\includegraphics[width=8cm]{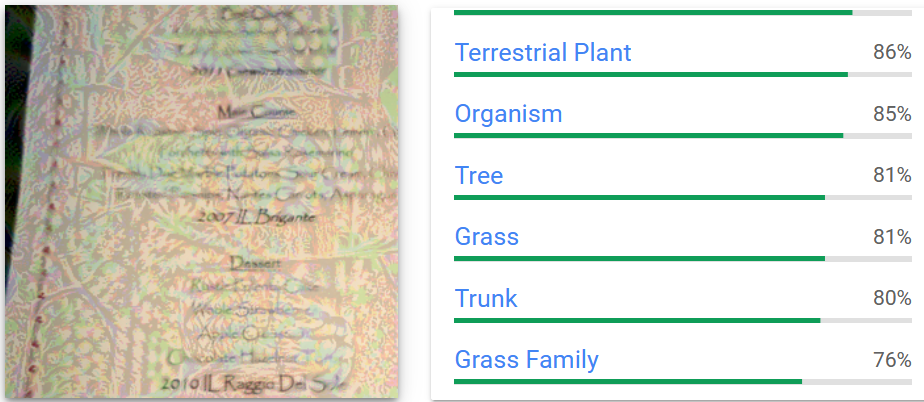}
    }
    \subfigure[]{
    	\includegraphics[width=8cm]{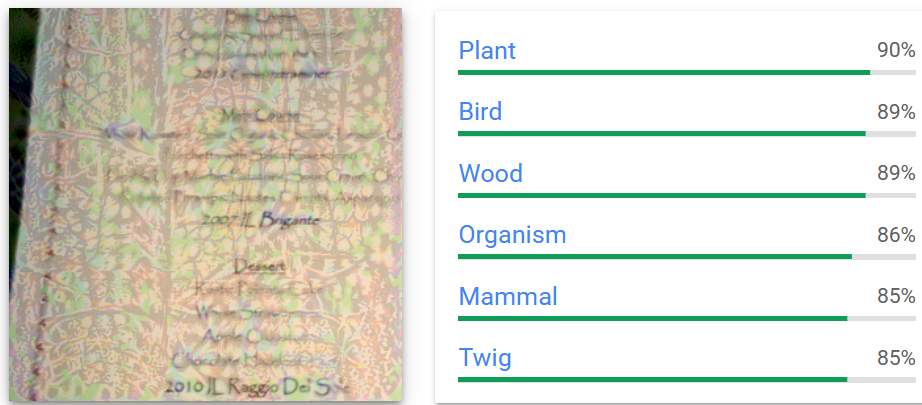}
    }
    \subfigure[]{
	    \includegraphics[width=8cm]{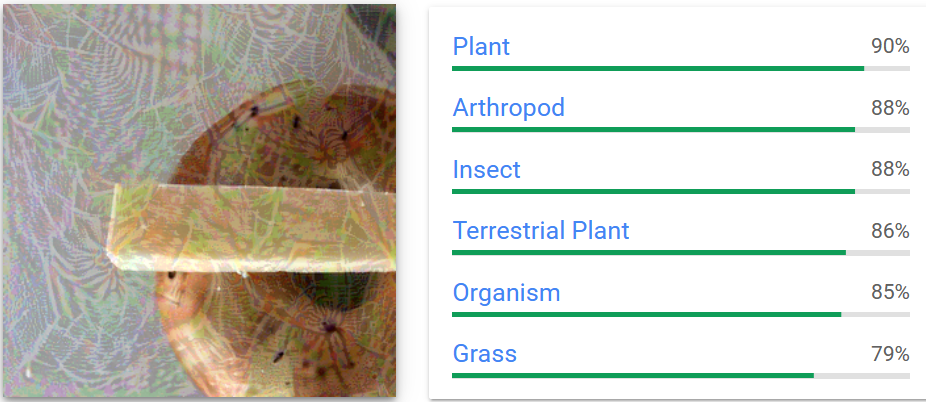}
    }
    \subfigure[]{
	    \includegraphics[width=8cm]{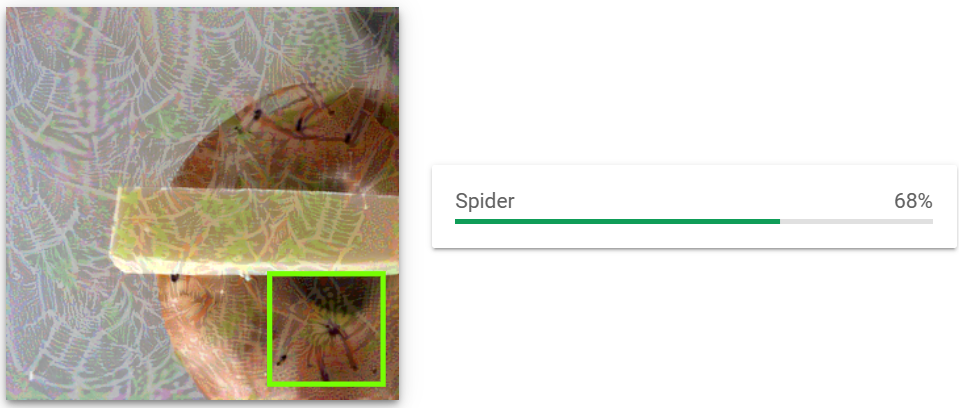}
    }
	\begin{center}
		Figure 2: AEs and the outputs from Google Cloud Vision API. The AEs are crafted against Res50adv with CFM (left) and the proposed CFM+\textit{everywhere} (right). From top to bottom, the target classes are `American lobster', `tench', `guacamole', `jay', and `black and gold garden spider' respectively.
	\end{center}
\end{figure*}

\end{document}